\documentclass[runningheads]{llncs}
\usepackage{multirow}
\usepackage{graphicx}
\usepackage{xcolor}
\usepackage{cite}
\usepackage{amsmath,amssymb,amsfonts}
\usepackage{algorithmic}
\usepackage{graphicx}
\usepackage{textcomp}
\usepackage{xcolor}
\begin{document}
\title{Improving the Results of \textit{De novo} Peptide Identification via Tandem Mass Spectrometry Using a Genetic Programming-based Scoring Function for Re-ranking Peptide-Spectrum Matches}

\titlerunning{Genetic Programming for Scoring Peptide-Spectrum Matches}
%

\authorrunning{S. Azari et al.}

\author{Samaneh Azari\inst{1}\and
	Bing Xue\inst{1} \and
	Mengjie~Zhang\inst{1}\and Lifeng~Peng\inst{2}}

\institute{School of Engineering and Computer Science, Victoria University of Wellington, \\ PO Box 600, Wellington 6140, New Zealand \\
	\email{\{samaneh.azari,bing.xue,mengjie.zhang\}@ecs.vuw.ac.nz}
	\and
	Centre for Biodiscovery and School of Biological Sciences, Victoria University of Wellington, PO Box 600,
	Wellington 6140, New Zealand
	\\
	\email{lifeng.peng@vuw.ac.nz} }
\maketitle  
\vspace{-5mm}
\begin{abstract}

\textit{De novo} peptide sequencing algorithms have been widely used in proteomics to analyse tandem mass spectra (MS/MS) and assign them to peptides, but quality-control methods to evaluate the confidence of \textit{de novo} peptide sequencing are lagging behind. A fundamental part of a quality-control method is the scoring function used to evaluate the quality of peptide-spectrum matches (PSMs). Here, we propose a genetic programming (GP) based method, called GP-PSM, to learn a PSM scoring function for improving the rate of confident peptide identification from MS/MS data. The GP method learns from thousands of MS/MS spectra. Important characteristics about goodness of the matches are extracted from the learning set and incorporated into the GP scoring functions. We compare GP-PSM with two methods including Support Vector Regression (SVR) and Random Forest (RF). The GP method along with RF and SVR, each is used for post-processing the results of peptide identification by PEAKS, a commonly used \textit{de novo} sequencing method. The results show that GP-PSM outperforms RF and SVR and discriminates accurately between correct and incorrect PSMs. It correctly assigns peptides to 10\% more spectra on an evaluation dataset containing 120 MS/MS spectra and decreases the false positive rate (FPR) of peptide identification. 
	
\end{abstract}
\vspace{-8mm}
\keywords{Genetic Programming  \and Symbolic Regression \and Peptide-spectrum Match \and Tandem Mass Spectrometry.}

\vspace{-4mm}

\section{Introduction}
\vspace{-2mm} 
 Mass spectrometry (MS) is the most commonly used method for the accurate mass determination and characterisation of proteins in complex biological samples. The common method for MS-based protein identification and characterising their amino acid sequences involves digesting proteins into peptides, which are then separated, fragmented, ionised, and captured by mass spectrometers. One of the common methods for assigning MS/MS spectra to peptide sequences is \textit{de novo} peptide sequencing which is particularly appropriate for discovering novel peptides which are not presented in any protein sequence database. Given a set of MS/MS spectra to a \textit{de novo} peptide sequencing algorithm, the results of peptide sequencing for each spectrum is a set of candidate peptides each having a confidence score indicating the quality of match between the spectrum and the candidate peptide. Normally, the highest-scoring (first ranked) candidate in each set of candidate peptides is regarded as the correct match for each spectrum. However, even with the identification of highest scoring PSMs, the fraction of peptide sequences that are fully correctly predicted by existing \textit{de novo} sequencing algorithms cannot achieve 70\%  \cite{yang2017psite}. The existence of noise, low quality of spectra, incomplete fragmentation and missing fragment ions could be possible reasons of incorrect full-length peptide sequencing. Therefore, the top-scored candidate does not necessary indicate a correct match and the correct match could be in the second or third rank in the candidate list. 

 As we do not want to assign a spectrum to a peptide which is not presented in the biological sample because incorrect peptide assignments result in incorrect protein identifications and the search scores in the current \textit{de novo} peptide sequencing algorithms do not always guarantee to find the true(correct) matches from many false matches, therefore, it is essential to apply a post-processing step as a PSM validation phase on the results of \textit{de novo} sequencing in order to improve peptide identification sensitivity and accuracy. Current \textit{de novo} peptide sequencing algorithms suffer from the lack of suitable scoring functions. The existing PSM-scoring functions to measure the goodness of a match between a spectrum and a peptide have the following limitations. A number of them are based on the simple shared peak count (SPC) approach where the number of peaks matched between experimental and theoretical (simulated) spectrum are counted \cite{colinge2007introduction} and the weight of all peaks are considered equal although some peaks are more informative than other peaks. Cross correlation based scores or statistical measures like the expectation value also have been previously used, but each one on its own does not serve as a strong discriminatory scoring function \cite{fenyo2003method}. In addition, some methods put a prior assumption and built a linear scoring function from combination of different similarity scores which measure the goodness of match between a spectrum and a peptide \cite{keller2002empirical}.
 
 Building a new scoring function from the possible combinations of different (sub)scores can be considered as a regression problem, where the (sub)scores are treated as features. Symbolic regression is a type of regression analysis that attempts to find the model that best fits a given dataset by discovering both model structure and parameters at the same time. Being a function identification process, symbolic regression does not face the problem of unknown gap in domain knowledge or human bias~\cite{babovic2000genetic,smits2005pareto}.	Having symbolic nature of solutions and being independent of any prior knowledge, GP is a promising method for symbolic regression problems. Symbolic regression using GP has been successfully applied to many real-world applications such as finance~\cite{ong2005building}, industrial processing~\cite{lee2011forecasting}, and software engineering ~\cite{harman2014search}. Therefore, it is worth discovering how GP employs different database search (sub)scores as its features and builds a regression model to reveal the intrinsic relationship of the data. The regression model will be used as the PSM-scoring function. The new scoring function will be used to re-score a collection of candidate PSMs resulting from \textit{de novo} peptide sequencing of MS/MS data. It is expected that the new GP-based function gives the highest score (first rank) to the correct match among other peptides belonging to the same candidate set, finding the correct peptide candidate for the given spectrum.
\vspace{-6mm}
\subsection{Research Goals}
\vspace{-2mm}
The main goal of this paper is developing an effective GP-based PSM scoring function to re-score and re-rank the PSMs which are the output of \textit{de novo} sequencing algorithms, aiming at improving the rate of full-length correct peptide identification of the \textit{de novo} algorithm. This problem will be formulated as a symbolic regression. 
The following \mbox{objectives} are specifically investigated:  
\begin{enumerate}
	\item Design appropriate terminal set, function set and a fitness function that help GP to explore the space of all possible combinations of similarity (sub)scores.
	\item Compare the performance of GP with other benchmark algorithms.  
	\item Evaluate the effectiveness of the new GP-based scoring function on the results of \textit{de novo} sequencing in terms of improvement in FDR.		 	
\end {enumerate}
	
\vspace{-5mm}
	\section{Background}
\vspace{-2mm}
	\subsection{Assigning MS/MS Spectra to Peptide Sequences}
	Basically, proteins and peptides are fundamentally the same as, being comprised of chains of amino acids that are held together by peptide bonds. 
	Peptide and protein identification is one of the significant challenges of proteomics. Mass spectrometry is the most commonly used techniques to overcome the challenge. A MS/MS spectrum is a mass to charge ratio plot which is the result of ionisation the biological sample by a mass spectrometer. The spectrum is used to identify the peptides in the sample and then from combining the peptides, proteins are identified. Mainly there are two main peptide identification strategy including database search and \textit{de novo} sequencing. A database search method matches the input experimental MS/MS spectra against the theoretical	spectra predicted for the peptides included in a protein sequence database search. However, when the protein database is not available or the biological sample contains unknown peptides and proteins \textit{de novo} sequencing methods are used. Peptide sequences are extracted directly from the MS/MS spectrum by measuring the mass differences between two informative peaks (b-/y-ions) corresponding to the mass of an amino acid and then linking the amino acids together to build a peptide.

	\subsection{Genetic Programming and MS/MS Data Analysis}
	Genetic Programming (GP) is a technique whereby a population of computer programs is evolved using an evolutionary algorithm to perform well in a predefined task. Randomly generating a set of individual as an initial population, GP searched for the solutions during the evolutionary search process. A set of genetic operators is applied on the individuals to generate fitter offsprings for the next generation \cite{langdon2008genetic}. GP simulates evolution by employing fitness based selection where the fittest program is expected to be chosen. The process can be stopped based on the stopping criteria which can be finding an ideal individual with a specified fitness
	value or reaching a maximum number of generations.
	
		\begin{figure*}[t]
		\centering
		\includegraphics [width=\columnwidth] {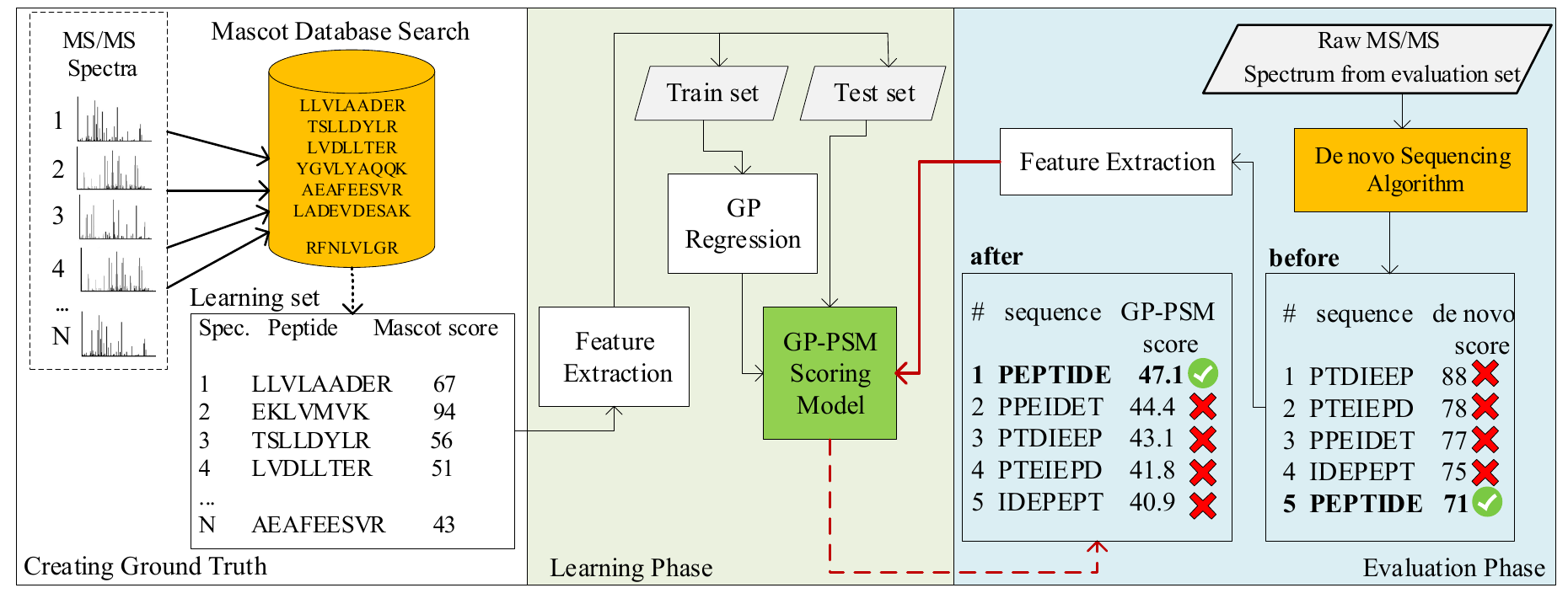}
		\vspace{-7mm}
		\caption{The workflow of the proposed GP-PSM method consisting of three phases.}
		\label{fig:flowchart}
	\end{figure*}

	\section{The Proposed Ranking GP Method}
	The proposed GP-PSM workflow designed for scoring the PSMs is presented in Fig. \ref{fig:flowchart}. The workflow proceeds in three phases: creating the ground truth, learning the GP-based PSM scoring function followed by an evaluation step to re-score the results of \textit{de novo} sequencing using the new scoring function. 
	
	Our ground truth is a set of MS/MS spectra with known identifications. The spectra used in the learning set are composed of various qualities with different peptide lengths varied from 5 to 12. More details about the learning set and validation set is given in Table \ref{table:dataset}. It is important to mention that the correct identification for each spectrum in both sets is known. This is an essential requirement since the existence of false positive PSMs in the learning set would not allow GP to find discriminating scoring functions. Therefore, here we have used a set of high confident PSMs identified and validated by Mascot database search, a benchmark database search tool. The Mascot database search has already identified the correct peptide corresponding to each spectrum with a Mascot score indicating the confidence level of the identification. 
	\color{black}
	
	For each instance in the learning set, a set of features explained in Table \ref{table:features} is extracted. The features are the (sub)scores or similarity scores which measure the quality of match between the experimental spectrum \textit{s} and the theoretical spectrum \textit{t}, the simulated spectrum of peptide \textit{p}, from different perspectives. After feature extraction, the instances from the learning set are divided into two sets train and test set (70\% and 30\%) to be used by GP to build the PSM scoring model. Therefore, the features are used as the independent variables and the Mascot score is the dependant variable in the GP regression model. GP uses train set to learn the model and applies the model on the test set. After generating the GP-PSM model, it is used to re-score the set of PSMs from the evaluation set. Each feature used in the model is relatively useful at determining if a PSM is correct or not, but the GP-PSM model incorporates them into a strong discriminatory scoring function for PSMs.
	 
	 After building the GP-PSM model, the effectiveness of this model is evaluated on the evaluation set and using the current \textit{de novo} sequencing algorithms. Given a raw MS/MS spectrum from the evaluation set to the \textit{de novo} sequencing algorithm, the result is a set of PSMs each having a \textit{de novo} score. The spectrum has been previously identified by Mascot database search and its corresponding peptide is known to us, but the \textit{de novo} sequencing algorithm does not know that and needs to assign a peptide to the spectrum. As it can be seen in the example, for the input spectrum, five candidate peptides are listed as the results of identification by the \textit{de novo} sequencing tool. The \textit{de novo} sequencing algorithm normally reports the highest-scoring candidate peptide as the results of the identification. As it can be seen in the example, the correct candidate is sit at the lowest score in the list (\textquoteleft before\textquoteright list). The scores of the PSMs in this list can be refined by using the GP-PSM model. So the first step is applying feature extraction and the second step is applying the GP model to re-score the PSMs. It is worth mentioning that the \textit{de novo} scores in the \textquoteleft before\textquoteright list are not used when the GP model is applied. As the result it can be seen that in the new re-scored list (\textquoteleft after\textquoteright list) the correct PSM got the highest score among the other candidates and this is what we would like to obtain.

	\color{black}
		
	\begin{table}[t]
		\caption{Features used in GP-PSM to represent a PSM.}
		\scriptsize
		\label{table:features}
		\vspace{-4mm}
		\centering
		\setlength{\tabcolsep}{8pt}
		\begin{tabular}{cll}
			\\[-2mm]\hline
			& \multicolumn{1}{c}{Feature name} & \multicolumn{1}{c}{Description} \\ \hline
			$ f_1 $ &$ I_{matched} $ & \begin{tabular}[c]{@{}l@{}}sum of intensities of matched peaks \end{tabular} \\ 
			$ f_2$ & $N_{matched}$ & \begin{tabular}[c]{@{}l@{}}\# of matched peaks\end{tabular} \\ 
			$ f_3$ & $N_{not-matched}$ & \begin{tabular}[c]{@{}l@{}}\# of un-matched peaks \end{tabular} \\
			$ f_4 $ & $\Delta$mass & \begin{tabular}[c]{@{}l@{}}The mass difference between the \textit{s} and \textit{p}\end{tabular} \\ 			
			$ f_5$ & Nterm & \begin{tabular}[c]{@{}l@{}}\# of matched b-ions from N-terminus\end{tabular} \\ 
			$ f_6 $ & Cterm & \begin{tabular}[c]{@{}l@{}}\# of matched y-ions from C-terminus\end{tabular} \\ 
			$ f_7 $ & Cos & Fixed length Normalised Dot product \\ 
			$ f_{8} $ & Euc & Fixed length normalised Euclidean distance \\ 
			$f_{9}$ & Hamming & Hamming distance between two vectorised \\ 
			$ f_{10} $ & SeqFix & Fixed length SEQUEST-like scoring  function \\ 
			$ f_{11} $ & SeqVar & Variable length SEQUEST-like scoring function \\ \hline
			
		\end{tabular}
	\end{table}

	\subsection{Feature Extraction}
 	For each PSM, a vector of 11 features summarised in Table \ref{table:features} is computed. These features
 	 measure the quality of match between the spectrum \textit{s} and the peptide \textit{p} based on different criteria. In order to match the experimental spectrum \textit{s} against the peptide \textit{p}, a theoretical spectrum \textit{t} is constructed from peptide \textit{p} based on the CID fragmentation rules \cite{herrmann2006peptide}. $ I_{matched} $ as the first feature in Table \ref{table:features}, calculates the sum of intensities of peaks matched between the experimental spectrum \textit{s} and the theoretical spectrum \textit{t}. $N_{matched}$ and $N_{not-matched}$ equals to the number of peaks in the theoretical spectrum \textit{t}, which are match and not matched against the spectrum \textit{s}, respectively. $\Delta$mass is the mass difference between the spectrum \textit{s} and the peptide \textit{p}. The $ Nterm $ feature counts the number of consecutive b-ions matched from N-terminus (left to right) of \textit{t} against \textit{s}. Similarly, the $ Cterm $ feature counts the number of y-ions from \textit{t} matched against \textit{s} from C-terminus side (right to left). The first six features were previously used as the fitness function of a genetic algorithm based \textit{de novo} sequencing \cite{GANovo2019} where all features were linearly combined with each feature having equal weights of 1. However, in this work we will not put the prior linear assumption on their combination and let GP to find the non-linear relationship between them.   
	
	Features $\{f_7,f_8,f_{9},f_{10},f_{11}\}$ vectorise the experimental spectrum \textit{s} and the theoretical spectrum \textit{t} into two binned vectors and then measure how well \textit{s} fits \textit{t}. Features $\{f_7,f_8,f_{9},f_{10}\}$ have fix length of 4,000, whereas $f_{11}$ has a variable length. The value in each bin in the vectorised experimental spectrum equals to the sum of the intensities of all peaks within the corresponding bin, whereas in the case of the vectorised theoretical spectrum the bin gets a value of one.
	The Cos feature, $\{f_7\}$ , uses dot/scalar product between two vectors \textit{s} and \textit{t} to calculate and normalises the result of the dot product to be in the range of [0,1] by dividing it into multiplication of the magnitude of the two vectors. A value of 0 for $f_7 $ indicates that the two spectrum vectors \textit{s} and \textit{t} have no peaks matched in between, whereas $f_7 =1 $ indicates a perfect match, and represents that  all peaks in \textit{t} are matched against those of \textit{s}. The Euc and Hamming features, $\{f_8,f_{9}\}$, calculate the normalised Euclidean distance and hamming distance between \textit{s} and \textit{t}, respectively.

	 The features SeqFix and SeqVar, ($f_{11},f_{12}$) are inspired from the  scoring function used in SEQUEST, a benchmark database search engine \cite{eng1994approach}. Both features apply a preprocessing step on the experimental spectrum in order to remove the potential noise peaks and normalise the intensities and then vectorise the spectra. SeqVar has a variable length for each spectrum and is determined by dividing the mass of the experimental spectrum into the fragment ion tolerance which here is 0.5. Both features use normalised dot product to measure the goodness of the match between \textit{s} and \textit{t}.
%
	\subsection{GP Program Representation}
	A tree based GP structure is considered to represent each GP individual in GP-PSM. Each individual represents a scoring function that returns a real number as the match score. A terminal set consisting of 11 features and random constants and a function set of arithmetic operators including $ \{+ , - , \times ,  /(\mbox{protected}) \} $ are considered for GP. A population size of 300 and maximum number of generations G = 100 are considered. The initial population is created based on ramped half-and half. The mutation and crossover rates are 0.1 and 0.9, respectively and the best individual is copied to the next generation. Tournament selection with size 5 selects the parental individuals. The algorithm is implemented in Python 3.6 and uses DEAP (Distributed Evolutionary Algorithms in Python) package \cite{DEAP_JMLR2012}.

	\subsection{An Effective Fitness Function for PSMs scoring}
	The GP method tries to generate a scoring function which combines different similarity (sub) scores as features and produces a real value score as the confidence score of the match between the experimental spectrum and the theoretical spectrum. The scoring function should be discriminating enough in order to distinguish a correct match from false matches in the evaluation phase (please see the flowchart in Fig. \ref{fig:flowchart}). As the GP problem is formulated as a symbolic regression task, we use relative sum of squared error (RSS) in Equation \ref{eq:rss} to compute the error of the prediction.  
	\vspace{-4mm}
		\begin{equation}
		RSS = \frac{\Sigma_{i=1}^{N}{(\hat{Y}_i - Y_i)}^2}{\Sigma_{i=1}^{N}{(\overline{Y} - Y_i)}^2} 
		\label{eq:rss}
		\vspace{-2mm}
		\end{equation}
	where $ \hat{Y}_i $ is the output of the GP individual corresponding to the target value $ Y_i $, $ \overline{Y} $ is the mean of the target values, and \textit{N} is the number of instances. A model with good performance has RSS$< $ 1. Therefore, in this problem GP tries to minimise the RSS.
 	\begin{table}[t]
 	\caption{The MS/MS spectra used in this study.}
 	\label{table:dataset}
 	\vspace{-7mm} 
 	\begin{center}
 		\begin{tabular}{|l|c|c|c|}
 			\hline
 			\multicolumn{2}{|c|}{dataset} & \# of MS/MS & \# of PSMs \\ \hline
 			\multirow{2}{*}{learning set} & train & 7,000 & 7,000 \\ \cline{2-4} 
 			& test & 3,000 & 3,000 \\ \hline
 			\multicolumn{2}{|l|}{evaluation set} & 120 & 600 \\ \hline
 		\end{tabular}
 	\end{center}
 \end{table}
	
	\section{Experiment Design}
	\subsection{MS/MS Datasets}
		
	To build the learning set and the evaluation set, the MS/MS spectra from the comprehensive full factorial LC-MS/MS benchmark dataset are used \cite{wessels2012comprehensive}. This dataset contains 50 protein samples extracted from \textit{Escherichia coli} K12 designed for evaluating MS/MS analysis tools. The MS/MS spectra are acquired from the linear ion trap Fourier-transform with the collision-induced dissociation (CID) technique. The peptide identification has been already applied by Wessels et al. \cite{wessels2012comprehensive} using Mascot v2.2 \cite{cottrell1999probability} with maximum missed cleavage of 1, precursor mass tolerance of 10 ppm, and fragment error of 0.8 Da with a cutoff q-value of 0.01. Therefore, the so called ground truth data (the peptide corresponding to each spectrum) is included in the full factorial dataset. 

	Since the fragmentation pattern strongly relies on the peptide\textquoteright \textit{s} charge and the precursor mass, this study only focuses on doubly charge peptides. The MS/MS spectra used in both learning and evaluation set have following characteristics: doubly charged, maximum precursor mass of 1150 Dalton, peptide length of 7 to 12 with no modifications. As it can be seen from Table \ref{table:dataset}, a set of 10,000 MS/MS spectra corresponding to 10,000 peptides are selected from the full factorial dataset to create the learning set. The learning set is split by 70\% and 30\% to create the train and test set, respectively. Also a set of 120 spectra as selected from the comprehensive dataset to create the evaluation set. These spectra later are given to a benchmark \textit{de novo} sequencing algorithm, called PEAKS \cite{ma2003peaks}, for peptide identification. For each spectrum, PEAKS produces at least five candidate peptides as the results of identification. That is the reason that the number of PSMs in evaluation set is not equal to the number of MS/MS spectra. The evaluation set is used for evaluating the effectiveness of GP-PSM in terms of improving the false discovery rate of the \textit{de novo} sequencing algorithm. More details is given in Experiments Section.


	\color{black}
	\subsection{Benchmark Algorithm}
	As GP is used to learn a scoring function in a regression task, the proposed method is compared with Random Forest (RF) and Support Vector Regression (SVR). Also as mentioned previously, the effectiveness of the scoring functions generated by GP, RF, SVR is evaluated by applying the model on the results of PEAKS. The results are evaluated in terms of false positive rate (FPR) before and after applying the model. Given the ground truth, FPR is the ratio of the number of correct matches to the total number of MS/MS spectra. 
	\begin{equation}
	FPR=\frac{FP}{N}
	\label{eq:FPR_PTM}
	\end{equation}
	Where $FP$ is the number of false-positive PSMs and $N$ indicates the number of MS/MS spectra. After applying the new GP-based scoring function on the results of PEAKS, in each group of five candidate peptides corresponding to the same spectrum, if the highest scoring peptide is not the correct match, the value of false positive increases by 1.    
	

	\subsection{Experiments}
	\subsubsection{Experiment I: Learning the PSM Scoring Functions}
	Based on the learning phase in the GP-PSM flowchart in Fig. \ref{fig:flowchart}, the three algorithms including GP, RF and SVR are used to learn the PSM scoring functions using the train set of the learning set from Table \ref{table:dataset}. The three algorithms are evaluated in terms of RSS measure in Equation \ref{eq:rss} on both train and test sets. As previously mentioned, the intention of the models is minimising the RSS value on both train and test sets. However, a method with low accuracy is
	sometimes superior to the one with high accuracy, therefore regardless of the performance of these three algorithms in learning phase, in evaluation phase each of them is applied on the evaluation set to investigate the effectiveness of each model as a post-processing method. 

	\subsubsection{Experiment II: Evaluating the Effectiveness of GP-PSM, RF and SVR}
	This experiment measures the performance of the three algorithms in terms of FPR before and after applying the models on the evaluation set. One question that might arise here is that what is the difference between the test set in the learning set and evaluation set. To answer this question we should explain how each set is created. The PSMs in the test set are the results of peptide identification by Mascot database search from the full factorial dataset. However, the PSMs in evaluation set are produced by PEAKS which is a \textit{de novo} sequencing algorithm. 
	
	Given an MS/MS spectrum to PEAKS, the output is a set of peptide sequences each having a confidence score between 0 and 100 \cite{ma2003peaks}. As previously mentioned even the highest-scored PSM does not necessary indicate a correct match for the corresponding spectrum, therefore the output of PEAKS is given as the input to the scoring functions generated by GP, RF and SVR for re-scoring. Then for each spectrum, the highest-scored PSM is reported as the final peptide identification. That is the reason that FPR before and after applying post-processing on the results of PEAKS for each algorithm is calculated. It is worth mentioning the already the correct peptide corresponding to the spectra in evaluation set is known, that is why FPR can be calculated.
	
	Another important point about the difference between the PSMs in the learning set and in the evaluation set is that as PEAKS reports five candidate peptide for each spectrum, the candidate are highly similar to each other, therefore, the value of their features also very close to each other. So the scoring function generated by the regression-based method should be strongly discriminated in order to give the highest score to the correct candidate peptide. 
	\color{black}

			\begin{table}[t]
		\centering
		
		\caption{The RSS results of the three methods in learning phase using the PSMs in learning set.}
		\vspace{-4mm} 
		\setlength{\tabcolsep}{12pt}
		\label{table:Learning_results}
		\begin{tabular}{lcc}
			\hline\\[-2mm]
			Method   & train  & test \\ \hline\\[-2mm]
			RF & 0.13 & 0.53 \\  
			SVR  & 0.69 & 0.67  \\ 
			GP-PSM & 0.55 $ \pm $ 0.04 & 0.55 $ \pm $ 0.03  \\
			Best GP individual & 0.49 & 0.50   \\
			
			\hline
		\end{tabular}
	\end{table}
\vspace{-4mm}
	\section{Results and Discussions}
	\vspace{-2mm}
	\subsection{Results of Experiment I:}
	Table \ref{table:Learning_results} presents the results of the experiment I when three methods including RF, SVR and GP were used to learn the PSM scoring function. The three methods are evaluated in terms of RSS on both train and test sets. For GP experiments, 30 individual runs using 30 different random seeds are considered.
	
	As it can be seen from Table \ref{table:Learning_results} that RF has the best result of train set and the second best is the Best individual of GP. On test set the best individual of GP has the best result. However, as it was previously mentioned from the results of learning set we might not be able to conclude which scoring function is the best, therefore, the three algorithm are applied on the evaluation set to check the effectiveness of the model. For GP the best individual in terms of RSS on train set is used at the PSM scoring function. More details are explained in the following section.  
	\begin{table}[t]
		\centering
		\caption{The results of PEAKS using the MS/MS spectra from evaluation set.}
		\label{table:PEAKS_result}
		\vspace{-3mm}
		\begin{tabular}{cc|c|c|c}
			\cline{3-4}
			\multicolumn{2}{c}{\multirow{2}{*}{}} & \multicolumn{2}{|c|}{FP} & \multirow{2}{*}{} \\ \cline{3-4}
			\multicolumn{2}{c}{} & \multicolumn{2}{|c|}{\# of MS/MS spectra that it's correct match} &  \\ \hline
			\multicolumn{1}{|c|}{Method} & \begin{tabular}[c]{@{}c@{}}TP\\ (Target PSMs)\end{tabular} & \begin{tabular}[c]{@{}c@{}}is not first-rank\\ (Missed Target PSMs)\end{tabular} & \begin{tabular}[c]{@{}c@{}}does not exist among the\\ five candidates\end{tabular} & \multicolumn{1}{c|}{FPR} \\ \hline
			\multicolumn{1}{|c|}{PEAKS} & 67 & 25 & 28 & \multicolumn{1}{c|}{0.44} \\ \hline
		\end{tabular}
	\end{table}
	
	\begin{table}[t]
		\caption{The results of PEAKS peptide identification after post-processing by RF, SVR and GP using the PSMs from evaluation set.}
		\label{table:evaluation_result}
		\vspace{-3mm}
		\centering
		\setlength{\tabcolsep}{1pt}
		\begin{tabular}{|l|c|c|c|c|c|c|}
			\hline
			Method & \begin{tabular}[c]{@{}c@{}}\# of Target PSMs \\ which are missed\\ (out of 67)\end{tabular} & \begin{tabular}[c]{@{}c@{}}\# of identified\\ Missed Target PSMs\\ (out of 25)\end{tabular} & TP & FP & FPR & \begin{tabular}[c]{@{}c@{}}FPR\\ reduction after\\ post-processing\end{tabular} \\ \hline
			RF & \textbf{6} & 7 & 68 & 52 & 0.43 & 1\% \\ \hline
			SVR & \textbf{6} & 11 & 72 & 48 & 0.40 & 4\% \\ \hline
			\textbf{GP-PSM} & 8 & \textbf{20} & \textbf{79} & \textbf{41} & \textbf{0.34} & \textbf{10\%} \\ \hline
		\end{tabular}
	\end{table}
	\subsection{Results of Experiment II:}
	This experiment is conducting in order to evaluate the performance of the new scoring functions generated by the three methods RF, SVR and GP on the evaluation set. PEAKS is used to perform \textit{de novo} sequencing using 120 MS/MS spectra from the evolution set. For each spectrum, PEAKS reports 5 candidate peptides each having a confidence score indicating the reliability of the match. In each set of candidates, the top-scored candidate is selected as the corresponding match for the input spectrum. Having the ground truth (from full factorial dataset), we can calculate the peptide identification rate. The results of peptide identification by PEAKS is shown in Table \ref{table:PEAKS_result}. It can be seen that out of 120 total number of spectra, only 67 of them are correctly identified by PEAKS (labelled as TP or Target PSMs in this table). However, there are 25 spectra that their corresponding correct peptide did not get the highest-rank in the candidate list. We call these PSMs as Missed Target PSMs which means they are those target PSMs which wrongly got lower scores by PEAKS. Therefore, other PSMs which were not supposed to get the highest scores belong to FP. Moreover, there are 28 spectra that non of the peptides in their candidate lists were correct. It can be seen that the FPR of PEAKS using evaluation set before applying the post-processing is 0.44. In overall, we expect the post-processing method to increase the number of TPs or Target PSMs, reduce the number of Missed Target PSMs and decrease the FPR. Please notice that for the set of 28 spectra, the post-processing method cannot help as the correct match does not exist among the candidate list of each spectrum.

	Table \ref{table:evaluation_result} presents the results of PEAKS \textit{de novo} peptide sequencing after post-processing by RF, SVR and GP using the PSMs from evaluation set which are the output of PEAKS. The second column in this table, \# of target PSMs which are missed, presents the number of target PSMs previously identified by PEAKS, but now are wrongly got low scores by the post-processing method. Also the third column in this table, number of identified missed target PSMs, indicates the number of missed target PSMs by PEAKS that are now identified by the post-processing method. 
	 The results show that the PSM scoring function generated by GP is able to identify 80\% ($ =(\frac{20}{25} \times 100) $) of the missed target PSMs whereas SVM and RF only found 44\% and 28\% , respectively. However, it can be seen that the RF and SVR are relatively better than GP in terms of keeping the target PSMs which are already got the highest ranks by PEAKS.  

	 Getting the lowest rate of FPR amongst the other methods, the GP method in overall outperforms other methods by 10\% reduction in FPR. The reason of the overall good performance of GP is its good discriminating ability. More analysis on the results of RF reveals that in most of failure cases assigned the similar rank to two candidate peptides in the list. As a rule in our method, if two PSMs belonging to the same candidate list if get the similar rank score, the identification is rejected even if one of them is the correct match.
	 In the case of SVR, quite often the target PSM did not have the change to get the highest score. This could be due to the low RSS results of SVR in the learning phase.

	\subsection{Analysis on the best GP evolved program}
\begin{figure}[t]
		\centering
		\includegraphics  [width =\columnwidth , height=7cm]  {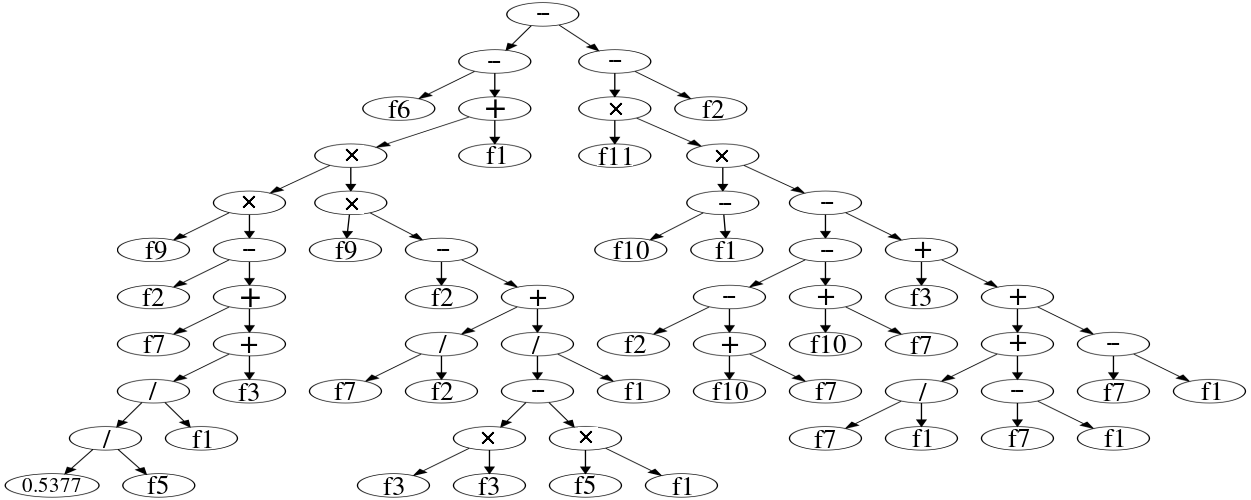}  
		\caption{The best GP evolved program (the PSM ranking function).}
		\label{fig:bestGP}
	\end{figure}

	Fig. \ref{fig:bestGP} shows the best GP-based scoring function among the 30 independent runs in terms of RSS on train set. The Tree contains 67 nodes. Having implicit feature selecting ability, GP automatically discarded irrelevant or redundant features such as $\{f_4\}$, $\Delta mass$, and $\{f_8\}$, Euclidean distance. These discards highly makes sense as for all the PSMs $\Delta mass$ is a small value and almost the same for all instances since the MS/MS spectra selected in this study do not have any post translation modifications so the mass difference between each spectrum and its corresponding peptide is very close to zero. As Cos feature and Euclidean features used here are both normalised and are mathematically equivalent and GP already has selected Cos feature, $\{f_7\}$, seven times so, $\{f_8\}$, Euc, is discarded by GP. As previously mentioned about finding the non-linear relationship between features $\{f_1\}$ to $\{f_6\}$, it can be seen that the left big sub-tree is mainly responsible for this task. It has 17 features and 13 of them are among features $\{f_1,f_2,f_{3},f_{4},f_{5}, f_6\}$, whereas the right big sub-tree is looking after the combination of the vectorised features including $\{f_7,f_8,f_{9},f_{10},f_{11}\}$.


	\section{Conclusions and Future Work}
	
	This work developed a genetic programming (GP) based method to automatically generate a PSM scoring function aiming at  reducing the rate of false discovery peptide identification from MS/MS data. The effective fitness function let GP to generate a strong discriminative scoring function which was able to improve the peptide identification. The GP method learns from thousands of MS/MS spectra. Important characteristics about goodness of the matches are extracted from the learning set and incorporated into the GP scoring functions. We compare GP-PSM with two methods including Support Vector Regression (SVR) and Random Forest (RF). The GP method along with RF and SVR, each is used for post-processing the results of peptide identification by PEAKS, a commonly used \textit{de novo} sequencing method. The results show that GP-PSM outperforms RF and SVR and discriminates accurately between correct and incorrect PSMs. It correctly assigns peptides to 10\% more spectra on an evaluation dataset containing 120 MS/MS spectra and decreases the false positive rate (FPR) of identification. 
 The results show that GP-PSM outperformed RF and SVR by 9\% and 4\% in terms of reduction the FPR, resulting in improving the PEAKS peptide identification . 
	Not being a black box, GP with its interpretability characteristic gives the chance to identify the important features which proven to be more informative for discriminating the correct PSMs from incorrect ones. The non-linear combination of the selected features can be used as the fitness function for other \textit{de novo} sequencing algorithms.

	As for future work, we will investigate how to design a wrapper GP method by considering the learning phase as the feature selection and the evaluation phase as the classification part of the wrapper method. We will also investigate how the new scoring function generated in this study can be used as a new feature for PSM scoring. 
\bibliographystyle{unsrt}
\bibliography{arxiv}

\end{document}